\pdfoutput=1

\documentclass[11pt]{article}

\usepackage{acl}

\usepackage{amsfonts}
\usepackage{amsmath}
\usepackage{amsthm}
\usepackage{adjustbox}
\usepackage{graphicx}
\usepackage{booktabs}
\usepackage{multirow}
\usepackage{multicol}
\usepackage{makecell}
\usepackage{times}
\usepackage{latexsym}

\usepackage[T1]{fontenc}

\usepackage[utf8]{inputenc}

\usepackage{microtype}

\usepackage{inconsolata}

\makeatletter
\renewcommand{\maketag@@@}[1]{\hbox{\m@th\normalsize\normalfont#1}}%
\makeatother

\title{The Integration of Semantic and Structural Knowledge \\ in Knowledge Graph Entity Typing   \vspace{-0.15cm}}

\author{
 \textbf{Muzhi Li\textsuperscript{1}},
 \textbf{ Minda Hu\textsuperscript{1}},
 \textbf{ Irwin King\textsuperscript{1}},
 \textbf{ Ho-fung Leung\textsuperscript{2}}
\\
 \textsuperscript{1}Dept. of Computer Science \& Engineering, The Chinese University of Hong Kong
 \\
 \textsuperscript{2}Independent Researcher
\\
\texttt{\{mzli,mindahu21,king\}@cse.cuhk.edu.hk} 
\\
\texttt{ho-fung.leung@outlook.com}
}

\begin{document}
\maketitle
\begin{abstract}
The Knowledge Graph Entity Typing~(KGET) task aims to predict missing type annotations for entities in knowledge graphs. Recent works only utilize the \textit{\textbf{structural knowledge}} in the local neighborhood of entities, disregarding \textit{\textbf{semantic knowledge}} in the textual representations of entities, relations, and types that are also crucial for type inference. Additionally, we observe that the interaction between semantic and structural knowledge can be utilized to address the false-negative problem. In this paper, we propose a novel \textbf{\underline{S}}emantic and \textbf{\underline{S}}tructure-aware KG \textbf{\underline{E}}ntity \textbf{\underline{T}}yping~{(SSET)} framework, which is composed of three modules. First, the \textit{Semantic Knowledge Encoding} module encodes factual knowledge in the KG with a Masked Entity Typing task. Then, the \textit{Structural Knowledge Aggregation} module aggregates knowledge from the multi-hop neighborhood of entities to infer missing types. Finally, the \textit{Unsupervised Type Re-ranking} module utilizes the inference results from the two models above to generate type predictions that are robust to false-negative samples. Extensive experiments show that SSET significantly outperforms existing state-of-the-art methods.~\footnote{\href{https://github.com/RaynorLEE/KG-EntityTyping}{https://github.com/RaynorLEE/KG-EntityTyping}}
\end{abstract}

\section{Introduction}

A knowledge graph (KG) is a graph-structured knowledge base that consists of triples in the form of $(\textit{head }\allowbreak \textit{entity}, \allowbreak\textit{relation}, \allowbreak\textit{tail }\allowbreak \textit{entity} )$. 
Each entity in the KG is annotated with one or multiple types. 
For example, 
entity ``$\textit{Albert Einstein}$'' belongs to the ``$\textit{/scientist/physicist}$'' type. Types (or \textit{concepts}) in KGs provide a high-level summary of their instance entities, which is crucial in many natural language processing~(NLP) tasks such as entity linking~\cite{EntLink}, relation extraction~\cite{RelExt}, and question answering~\cite{ijcai2022p427}.

Currently, most KGs (e.g., Freebase~\cite{Freebase} and YAGO~\cite{YAGO}) are manually constructed by domain experts, which the type annotations are usually incomplete. The statistics on the FB15kET dataset show that 10\% of the entities with the type ``\textit{/music/artist}'' do not have type  ``\textit{/people/person}''~\cite{ETE}. In light of this, we focus on the Knowledge Graph Entity Typing~(KGET) task, which aims to infer missing type annotations for entities in a KG.

\begin{figure}
    \vspace{-0.35cm}
    \centering
    \includegraphics[width=0.49\textwidth]{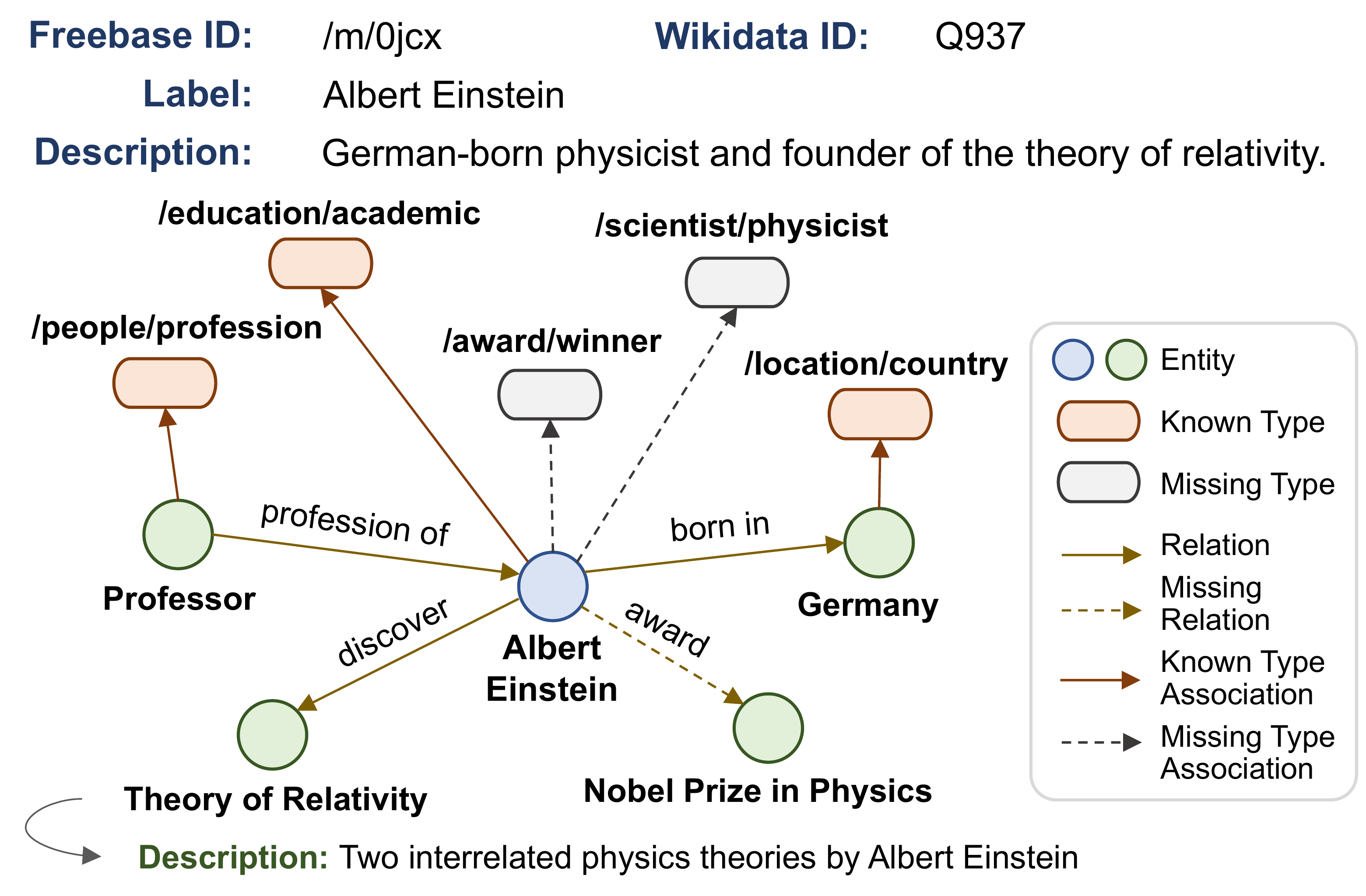}
    \caption{A knowledge graph fragment centered on entity ``\textit{Albert Einstein}". We aim to infer missing types of the target entity based on the structural and textual information provided in the local subgraph.}
    \label{fig:kg}
    \vspace{-0.45cm}
\end{figure}


Recent studies in the KGET task can be categorized into three groups: (1) embedding-based methods~\cite{ETE,ConnectE}, (2) GNN-based methods~\cite{MRGAT,AttEt,CET,Miner}, and (3) transformer-based methods~\cite{TET}. However, each group of solutions has its drawbacks. 
Specifically, embedding-based methods ignore the rich neighbor information of entities. GNN-based methods fall short of capturing the interactions between non-directly connected neighbors of the target entity~\cite{TET}. The only transformer-based method TET suffers from its high computational complexity. More importantly, all existing methods \textit{ignore the textual information in the KG} and \textit{suffer from serious false-negative problem}~\cite{Miner}, which ultimately limits the type inference performance. 

We argue that \textit{the textual representations of entities, relations, and types provide important semantic knowledge for type inference.} 
For example, in Figure~\ref{fig:kg}, it is hard to infer that ``\textit{Albert Einstein}'' belongs to the type  ``\textit{/scientist/physicist}'' based solely upon the structural knowledge since triple (\textit{Albert Einstein}, \textit{award}, \textit{Nobel Prize in Physics}) is missing from the KG. Fortunately, the description of the neighbor entity ``\textit{Theory of Relativity}'' reveals that the theory is in the domain of ``\textit{physics}'', which can help us to conclude that ``\textit{Albert Einstein}'' has ``\textit{/scientist/physicist}'' type. 

 Moreover, we observe that \textit{semantic and structural knowledge can complement each other to alleviate the false-negative problem. }
Some valid type annotations are plausible but missing from the KG. 
Simply considering these missing type annotations as negative samples will result in significant performance degradation. 
We find that the semantic knowledge obtained from the textual information provides an informative prior for the alleviation of the false-negative problem. If both semantic and structural knowledge support that an entity should belong to a particular type, we can assign a higher relevance score to that type. 
Such an agreement allows us to mitigate the impact of false-negative samples with high confidence. In summary, the main objective of this paper is to capture and integrate semantic and structural knowledge from the local subgraph of each entity in the KG and apply both types of knowledge to the entity typing task.

Based on the aforementioned observations, we propose a novel \textbf{\underline{S}}emantic and \textbf{\underline{S}}tructure-aware KG \textbf{\underline{E}}ntity \textbf{\underline{T}}yping~{(SSET)} framework, which consists of three modules. The \textit{Semantic Knowledge Encoding module} encodes factual knowledge in the KG with a Masked Entity Typing task. 
The \textit{Structural Knowledge Aggregation module} leverages knowledge from $1$-hop neighbors, multi-hop neighbors, and known types of entities to predict missing types. The \textit{Unsupervised Type Re-ranking module} utilizes the preliminary inference results obtained from the two modules mentioned above to generate type predictions that are robust to false-negative samples.
Our contributions are summarized as follows:
\begin{itemize}
    \setlength\itemsep{0.1em}
    \item We propose a novel framework SSET which encodes and utilizes semantic and structural knowledge for the KGET task. 
    
    \item We notice the complementary relationship between semantic and structural knowledge and integrate the two types of knowledge to alleviate the impact of false-negative samples.
   
    \item We conduct empirical and ablation experiments on two widely used datasets, showing that SSET significantly outperforms the existing state-of-the-art approaches.
\end{itemize}


\section{Related Work}
\paragraph{Embedding-based Methods.} 
ETE~\cite{ETE} is the first model specialized for the KGET task. It employs KG embedding methods~\cite{TransE,ContE} to embed entities and relations, and learns type embeddings by minimizing the distance between embeddings of entities and corresponding types. ConnectE~\cite{ConnectE} jointly embeds entities and types into different spaces and proposes two type inference mechanisms, namely ``E2T'' and ``TRT''. Despite the simplicity and intuitiveness, these methods ignore the rich neighbor information of entities, which seriously limits their performance~\cite{Miner}.

\paragraph{Graph Neural Network-based Methods.}
Graph neural networks (GNNs)~\cite{GCN,SSNC} have been proven to be effective in capturing structural knowledge from graph-structured data, including KGs. Recently, several GNNs such as RGCN~\cite{RGCN}, ConnectE-MRGAT~\cite{MRGAT} have been proposed to better aggregate neighbor information in KGs and are applied in the KGET task. Most notably, CET~\cite{CET} leverages knowledge from each neighbor of an entity in an independent and aggregated manner to mitigate the impact of noises from irrelevant neighbors. MiNer~\cite{Miner} further extends the scope of knowledge aggregation to multi-hop neighbors. 
Nevertheless, these methods primarily focus on modeling the graph structure and overlook the textual information provided in KGs. 

\paragraph{Transformer-based Methods.} To the best of our knowledge, TET~\cite{TET} is the only transformer-based method specified for the KGET task. TET utilizes three transformers to encode the neighbors of each entity and model the knowledge interactions between different neighbors. However, these transformers have no relationship with pre-trained language models (PLMs)~\cite{BERT,su-etal-2022-roberta}, which cannot model semantic knowledge in the textual representations of entities, relations, and types. 

\paragraph{Semantic-aware Fine-grained Entity Typing. } Recently, several methods have been proposed to incorporate semantic information into the fine-grained entity typing (FET) task~\cite{HMGCN}. Notably, Cat2Type~\cite{Cat2Type} leverages Wikipedia category names to embed entity mentions for type prediction. \citet{GraphWalk} further employs graph walks to aggregate neighborhood information of entities from an external knowledge base. It should be noted that the FET task focuses on inferring the semantic type of an entity mention within plain text. Due to the differences in problem settings, it is infeasible to directly apply these FET solutions to the KGET task. 



\section{Problem Specification}
We consider a KG $\mathcal{G}$ to consist of triples in the form of $(e_s, r, e_o)$ and entity type assertions in the form of $(e, t)$, where $e_s, e_o, e \in \mathcal{E}$, $r \in \mathcal{R}$, $t \in \mathcal{T}$. The notations $e_s$ and $e_o$ denote the subject and the object of a triple. $\mathcal{E}, \mathcal{R}, \mathcal{T}$ are the set of entities, relations, and types, respectively. In KGs, an entity can have multiple types, but some type annotations are missing from the KG. Here, we use $\mathcal{T}(e)=\{t|(e,t)\in \mathcal{G}\}$ to denote the set of known types of entity $e$. We denote the set of triples related to entity $e$ by $\mathcal{N}(e)=\{(e, r, \tilde{e}) \cup (\tilde{e}, r, e) \in \mathcal{G}\}$. In most KG datasets such as FB15k~\cite{Freebase} and YAGO43k~\cite{YAGO}, entities are provided with \textit{labels} and \textit{descriptions}, while relations and types are represented by their \textit{textual identifiers}. We assume that these \textit{textual information} are meaningful and contain valuable semantic knowledge for the KGET task. In this paper, we aim to reveal missing type annotations based on the local graph structure of entities and corresponding textual information. 
\begin{figure*}
    \centering
    \includegraphics[width=\textwidth]{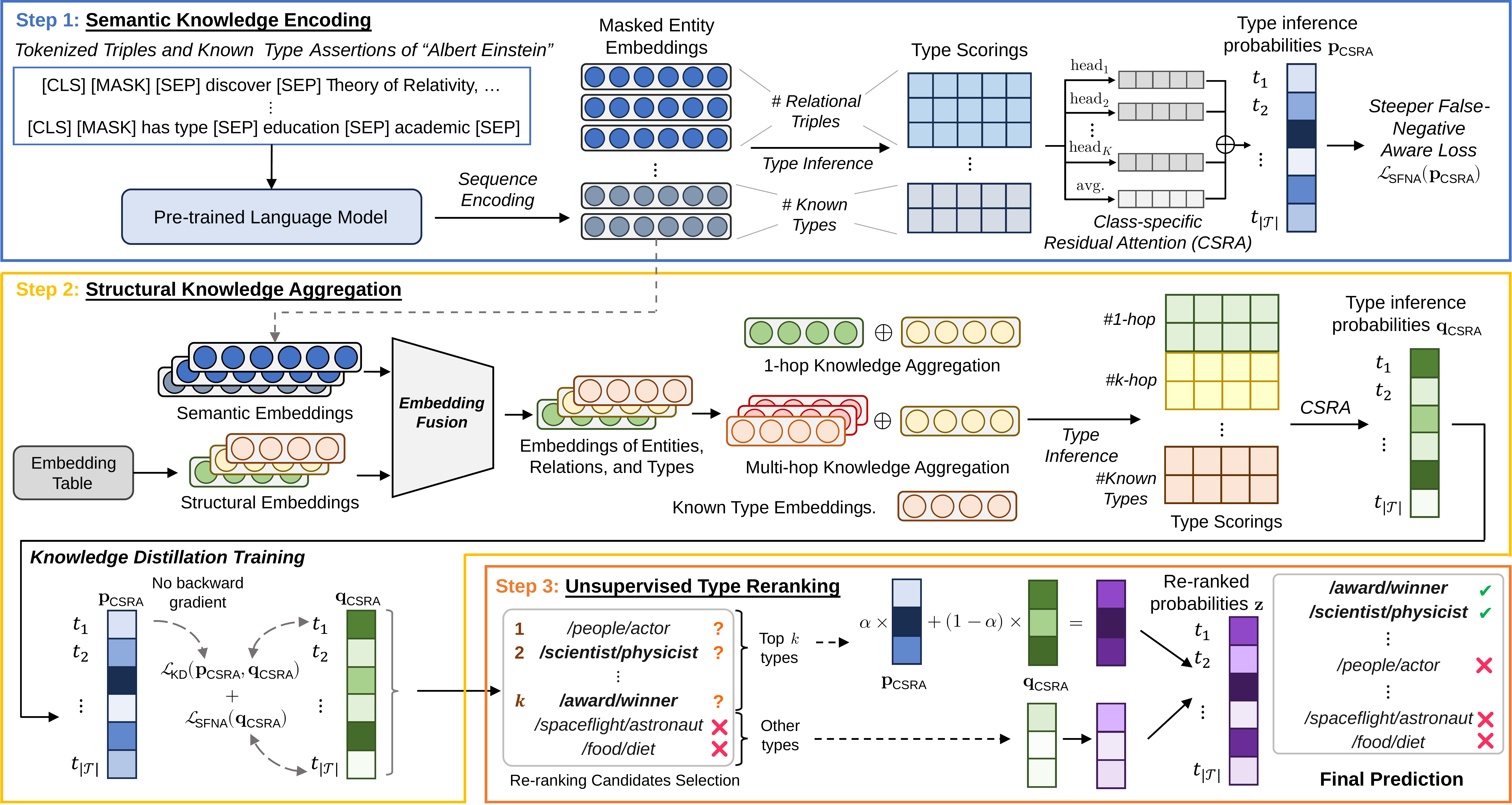}
    \caption{The end-to-end architecture of the SSET, consists of three modules: Semantic Knowledge Encoding module (top), Structural Knowledge Aggregation module (mid), and Unsupervised Type Re-Ranking module (bot). }
    \label{model_figure}
    \vspace{-0.20cm}
\end{figure*}

\section{Method}
Figure~\ref{model_figure} provides an overview of the architecture of our proposed SSET framework, which consists of three modules. First, the \textit{1) Semantic Knowledge Encoding module}~(SEM) infuses the knowledge from triples and entity-type assertions into the pre-trained language model~(PLM). We expect the fine-tuned language model can be utilized to encode textual representations of entities, relations, and types, and to conduct preliminary type inference. Then the \textit{2) Structural knowledge aggregation module}~{(SKA)} aggregates knowledge from the local subgraph of each target entity and utilizes $1$-hop neighbors, multi-hop neighbors, and known types of the entity to infer its missings types. Finally, the \textit{3) Unsupervised Type Re-ranking module~(UTR)} exploits the agreement between semantic and structural knowledge to alleviate the false-negative problem and generate the final type rankings.


\subsection{Step 1: Semantic Knowledge Encoding}
To incorporate the factual knowledge from the KG into the PLM, we propose a fine-tuning task known as Masked Entity Typing~(MET). The MET task recovers missing types of the masked entity by utilizing \textit{triples} and \textit{entity-type assertions} that contain the entity. Since triples and entity-type assertions in KGs are differ from natural language sentences, we will first introduce the tokenization process before delving into the details of the MET task.

\subsubsection{Tokenization}
Before being inputted into the PLM encoder, triples and entity-type assertions need to be serialized into token sequences. Formally, we tokenize each triple $(e_s,r,e_o)$ into the following sequence:
\vspace{-0.15cm}
\begin{equation}
\resizebox{.86\linewidth}{!}{$
    [\text{CLS}]\text{ }\text{text}_{e_s}\text{ }[\text{SEP}]\text{ }\text{text}_r\text{ }[\text{SEP}]\text{ }\text{text}_{e_o} \text{ }[\text{SEP}], $}
    \vspace{-0.1cm}
\end{equation}
where $\text{text}_e = [\text{label}_e, \text{desc}_e]$ contains the label and the description of entity $e$, ``$\text{text}_r$'' is the textual identifier of relation $r$, $e_s$ and $e_o$ differentiates the subject and object of a triple. 
Similarly, we tokenize each entity-type assertion as follows:
\vspace{-0.1cm}
\begin{equation}
\resizebox{.895\linewidth}{!}{$
    [\text{CLS}]\text{ }\text{text}_e\text{ }[\text{SEP}]\textit{ has type }[\text{SEP}]\text{ }\text{text}_t\text{ }[\text{SEP}], $}
\vspace{-0.1cm}
\end{equation}
where ``\textit{has type}'' is a predicate connecting an entity and a type, ``$\text{text}_t$'' is the textual identifier of type~$t$. We observed that types within some KGs such as FB15k~\cite{Freebase} usually have multiple levels (e.g., ``\textit{/scientist/physicist}''). During the tokenization process, different levels of a type will be separated with a [SEP] token (e.g., ``\textit{scientist} [SEP] \textit{physicist}''). 

\subsubsection{Masked Entity Typing~(MET) Task}
In KGs, the triples and known types of an entity contribute differently to the inference of a specific missing type. For example, the triple (\textit{Albert Einstein, award, Nobel Prize}) indicates that the entity ``Albert Einstein'' has the type ``/award/winner''. However, the triple (\textit{Albert Einstein, born in, Germany}) does not provide relevant support for the same conclusion. To mitigate the impact of irrelevant information, we separately encode the token sequences of each triple and entity-type assertion that involves the target entity.

To prevent the PLM from overfitting to the label and description of the target entity, we mask the tokens that correspond to the target entity in all relevant token sequences mentioned above. This operation forces the PLM to capture semantic knowledge from the textual representations of relations, neighboring entities, and known types. 
Subsequently, the masked sequences are inputted into the language model~(LM) encoder~\cite{BERT,su-etal-2022-roberta}.\footnote{We utilize pre-trained BERT and RoBERTa encoder as the backbone of the SEM module due to their simplicity. The proposed SSET framework can be generalized to any PLMs that can be fine-tuned such as GPT.}  Finally, we apply a non-linear classifier to the embeddings of the [MASK] tokens obtained from the output of the LM encoder:
\vspace{-0.1cm}
\begin{equation}\label{classifier}
    \mathbf{S} = \mathbf{W}\sigma(\mathbf{H}^{\text{mask}})+\mathbf{b},
    \vspace{-0.1cm}
\end{equation}
where $\mathbf{W}\in\mathbb{R}^{|\mathcal{T}|\times d}$ and $\mathbf{b} \in \mathbb{R}^d$ are the learnable parameters. $\mathbf{H}^{\text{mask}} = [\mathbf{h}_1^{\text{mask}}, \mathbf{h}_2^{\text{mask}}, \cdots,\mathbf{h}_{m+n}^{\text{mask}}] \in \mathbb{R}^{(m+n)\times d}$ are the [MASK] token embeddings from the $m$ triples and $n$ known types of the target entity. $\sigma(\cdot)$ denotes the ELU activation function. The output $\mathbf{S}=[\mathbf{s}_1, \mathbf{s}_2, \cdots, \mathbf{s}_{m+n}] \in \mathbb{R}^{(m+n)\times{|\mathcal{T}|}}$  consists of type scoring vectors for the $m$ triples and $n$ known types. In particular, $s_i^j$ represents the relevance score between the $i$-th triple (or known type) of the entity and the $j$-th type in the KG.

Following the state-of-the-art KGET model MiNer~\cite{Miner}, we adopt the class-specific residual attention mechanism (CSRA)~\cite{CSRA} to generate the preliminary type inference result.  CSRA suggests utilizing $H$ different temperatures $T_1, \cdots, T_H$ to pool the type relevance scores of the $m$ triples or $n$ known types. This process yields preliminary inference probabilities $\mathbf{p}_{\text{CSRA}} = [p^1, p^2, \cdots, p^{|\mathcal{T}|}]$ that the target entity belongs to each of the $|\mathcal{T}|$ types:
\vspace{-0.1cm}
\begin{equation}
    \mathbf{p}_{\text{CSRA}} = \text{sigmoid}(\sum_{h=1}^H (\mathbf{s}_{T_h} + \frac{1}{m+n}\sum_{i=1}^{m+n}\mathbf{s}_i)),
\end{equation}
where $\mathbf{s}_{T} = [s_{T}^1, \cdots, s_{T}^{|\mathcal{T}|}] \in \mathbb{R}^{|\mathcal{T}|}$ is the pooled type scoring vector under temperature $T$. Here, we have
\begin{equation}
    s_T^j = \sum_{i=1}^{m+n}\frac{\exp{(Ts_i^j)}}{\sum_{k=1}^{m+n}\exp{(Ts_k^j)}}\cdot s_i^j.
\end{equation}
We utilize the steeper false-negative aware (SFNA) loss~\cite{TET} to fine-tune the PLM:
\begin{align}
    \mathcal{L}_{\text{SFNA}}(\mathbf{p}_{\text{CSRA}}) = &\sum_{(e_i, t_j) \notin \mathcal{G}}f(p_i^j)\log (1-p_i^j) \notag
    \\&- \sum_{(e_i, t_j) \in \mathcal{G}} \log (p_i^j),
    \vspace{-0.05cm}
\end{align}
where $p_i^j$ is the probability that entity $i$ belongs to type $j$, $f(\cdot)$ is a re-weighting function~(see Eq.~(\ref{reweighting})) that penalizes negative samples with too high or too low probabilities. 

\subsection{Step 2: Structural Knowledge Aggregation}
In KGs, an entity's knowledge is manifested within its local subgraph. Simply considering the one-hop neighbors and known types of an entity is insufficient to reveal all of its missing types. The inference of certain challenging types necessitates a comprehensive understanding derived from the multi-hop neighbors of the target entity. Hence, in this module, we aim to utilize \textit{$1$-hop neighbors}, \textit{multi-hop neighbors}, and the \textit{known types} of the target entity to predict missing types.~\footnote{We do not treat types of an entity as its neighbors. Distinguishing entity types from entities helps us to achieve better results.}

\subsubsection{Embedding Fusion}
Ideally, multi-hop structural knowledge improves the performance of type inference. However, without good initial embeddings, the inference results will be unreliable. In such cases, the SKA module cannot receive meaningful gradients to update the parameters and may learn to disregard the structural knowledge~\cite{Ziniu}. 

To overcome the ``cold-start'' problem, we embed and cache textual representations of entities, relations, and types (denoted as $\mathbf{h}_e^{(t)}$, $\mathbf{h}_r^{(t)}$, and $\mathbf{h}_t^{(t)}$) from the fine-tuned LM. These embeddings are fixed during the training process, as the textual representations are independent of the KG's connectivity. In addition, to enhance the scalability of our model, we also incorporate a set of learnable structural embeddings for each entity, relation, and type, which are denoted as $\mathbf{h}_e^{(s)}$, $\mathbf{h}_r^{(s)}$, and $\mathbf{h}_t^{(s)}$. 

Before knowledge aggregation, we fuse the textual and structural embeddings of each entity, relation, and type into a unified latent space. This is achieved by two multi-layer perceptions (MLPs) followed by a $\mathcal{L}_2$ normalization. Formally, we have
\vspace{-0.08cm}
\begin{equation}~\label{embfuse}
\small
    \mathbf{h}_{\{\cdot\}} = \frac{\text{MLP}_t(\mathbf{h}_{\{\cdot\}}^{(t)})}{{||\text{MLP}_t(\mathbf{h}_{\{\cdot\}}^{(t)})||}_2} + \frac{\text{MLP}_s(\mathbf{h}_{\{\cdot\}}^{(s)})}{{||\text{MLP}_s(\mathbf{h}_{\{\cdot\}}^{(s)})||}_2},
    \vspace{-0.03cm}
\end{equation}
where $\mathbf{h}_{\{\cdot\}}$ is the \textit{unified knowledge embedding} of an entity, a relation, or a type. For simplicity, we denote $\mathbf{h}_{e}, \mathbf{h}_{r}, \text{ and } \mathbf{h}_{t}$ as $\mathbf{e}, \mathbf{r}, \text{ and } \mathbf{t}$ hereinafter. 

\subsubsection{One-hop Knowledge Aggregation} 
Following the convention adopted in CET~\cite{CET} and MiNer~\cite{Miner}, we utilize TransE~\cite{TransE} to aggregate the knowledge from each of the triples containing the target entity. Formally, for target entity $e$, its representation aggregated from the one-hop neighbor entity $\tilde{e}$ and the relation $r$ can be computed as:
\vspace{-0.1cm}
\begin{equation}~\label{onehopagg}
    \mathbf{h}_{(r, \tilde{e})}^{(1)}=(\tilde{\mathbf{e}} - \mathbf{r}),
\vspace{-0.1cm}
\end{equation}
where $\tilde{\mathbf{e}}$ and $\mathbf{r}$ are the embeddings of entity $\tilde{e}$ and relation $r$, $(e, r, \tilde{e})$ is a triple in the KG.\footnote{We also consider the neighbor entities $\tilde{e}$ connected by reversed relations $r^{-1}$ (i.e. $(\tilde{e},r^{-1},e) \in \mathcal{G}$). We embed reversed relations by the negation of the embedding of the original relation s.t. $\mathbf{r}^{-1} = -\mathbf{r}$.}  

\subsubsection{Multi-hop Knowledge Aggregation} 
Now, we introduce how we aggregate structural knowledge from the multi-hop neighborhood of the target entity. Inspired by MiNer~\cite{Miner}, we iteratively represent each $(k-1)$-hop neighbor entity by its known types and related $k$-hop entities. We use $2$-hop neighbors as an example to illustrate the knowledge aggregation process and will generalize to the case of more hops. 

Considering the target entity $e$, we encode the $2$-hop structural knowledge via each $1$-hop neighbor connected to the entity:
\vspace{-0.1cm}
\begin{equation}
    \mathbf{h}_{(r,\tilde{e})}^{(2)} = f_\text{agg}^{(1)}(\tilde{e}) - \mathbf{r},
    \vspace{-0.1cm}
\end{equation}
where $\mathbf{h}_{(r,\tilde{e})}^{(2)}$ is the representation of entity $e$ aggregated from the $2$-hop neighborhood via the $1$-hop neighbor $\tilde{e}$. $f_{\text{agg}}^{(1)}$ is an aggregation function that computes the representation of an entity with its related entities and known types:
\begin{equation}
\resizebox{.875\linewidth}{!}{$
    \displaystyle
    f_{\text{agg}}^{(1)}(e) = \frac{\sum_{(e,r,\tilde{e}) \in \mathcal{N}(e)}{(\tilde{\mathbf{e}} -\mathbf{r})} + \sum_{t\in \mathcal{T}(e)} \mathbf{t}}{|\mathcal{N}(e)| + |\mathcal{T}(e)|}, $}
\end{equation}
where $\mathcal{N}(e)$ is the set of triples containing entity $e$, $\mathcal{T}(e)$ is the set of entity $e$'s known types, $\mathbf{t}$ is the embedding of type $t$. 
We can iteratively generalize the knowledge aggregation process to multi-hop neighbors $(k \geq 3)$ of the target entity with the following two equations:
\vspace{-0.1cm}
\begin{equation}
    \mathbf{h}_{(r,\tilde{e})}^{(k)} = f_\text{agg}^{(k-1)}(\tilde{e}) - \mathbf{r},
    \vspace{-0.1cm}
\end{equation}
\begin{equation}
\resizebox{.99\linewidth}{!}{$
    \displaystyle
    f_{\text{agg}}^{(k)}(e) = \frac{\sum_{(e,r,\tilde{e}) \in \mathcal{N}(e)}{(f_{\text{agg}}^{(k-1)}(\tilde{e}) -\mathbf{r})} + \sum_{t\in \mathcal{T}(e)} \mathbf{t}}{|\mathcal{N}(e)| + |\mathcal{T}(e)|}. $}
    \vspace{-0.1cm}
\end{equation}

\subsubsection{Type Inference}
We construct a matrix $\mathbf{H}^{\text{agg}} = [\mathbf{h}_{(r_1,\tilde{e}_1)}^{(1)},\cdots,\allowbreak\mathbf{h}_{(r_m,\allowbreak\tilde{e}_m)}^{(K)},\allowbreak\mathbf{t}_1,\cdots,\mathbf{t}_n]$ by concatenating the entity representations aggregated from the $1,2,\cdots,K$-hop neighborhood and the embeddings of each known type. Similar to the SEM module, we apply a non-linear classifier to the embedding matrix and utilize the CSRA~\cite{CSRA} mechanism to compute the probabilities $\mathbf{q}_{\text{CSRA}} = [q^1,\allowbreak q^2, \cdots, q^{|\mathcal{T}|}]$ that the entity belongs to each type.

\subsubsection{Knowledge Distillation} 
Some plausible type annotations are missing from the KG. Simply considering missing types as negative samples will result in a serious false negative problem. 
We find that the preliminary type inference results obtained from the SEM module provide an informative prior that can be leveraged 
to alleviate the impact of those false-negative samples. 
Hence, we add the following knowledge distillation~(KD) loss to align the inference results obtained from the SEM and the SKA modules:
\vspace{6pt}
\vspace{-0.15cm}
\begin{align}
    \mathcal{L}_{\text{KD}} (p, q)= &\sum_{i=1}^{|\mathcal{E}|}\sum_{j=1}^{|\mathcal{T}|}(1-p_i^j)f(q_i^j)\log(1-q_i^j) \notag
    \\ & -\sum_{i=1}^{|\mathcal{E}|}\sum_{j=1}^{|\mathcal{T}|} p_i^j\log(q_i^j),
    \vspace{1pt}
    \vspace{-0.15cm}
\end{align}
where $p_i^j, q_i^j$ are the probabilities that entity $i$ belongs to type $j$ obtained from the two modules, respectively. $f(\cdot)$ is the re-weighting function defined in Eq.~{(\ref{reweighting})}.
The training objective for the SKA module is a weighted average of the KD loss and the SFNA loss:
\vspace{-0.05cm}
\begin{align}
    \mathcal{L}_{\text{SKA}} =&\text{ }\lambda \cdot \mathcal{L}_{\text{SFNA}}(\mathbf{q}_{\text{CSRA}}) \notag
    \\& + (1-\lambda) \cdot \mathcal{L}_{\text{KD}} (\mathbf{p}_{\text{CSRA}}, \mathbf{q}_{\text{CSRA}}),
\vspace{-0.05cm}
\end{align}
where $\lambda \in [0, 1]$ is a hyperparameter balancing the two loss functions.

\subsection{Step 3: Unsupervised Type Re-ranking}
During the inference stage, the SEM and SKA modules generate two sets of probabilities to assess the likelihood of each entity belonging to each type in the KG. Motivated by the complementary nature of semantic and structural knowledge, we leverage the agreement between the probability outputs of both modules to generate our final type predictions.  

Specifically, if both types of knowledge support that an entity belongs to a particular type, we will simultaneously obtain high probabilities from both modules. In such cases, we assign a lower (better) rank to the type with confidence. Conversely, if either module generates a low probability for a type, we assign it a higher rank. To achieve the aforementioned goals, we employ the re-ranking mechanism as in~\cite{RobustKGC}. Specifically, for each entity $i$, we retrieve the top $k$ candidate types with the highest probabilities obtained from the SKA module. Then we re-evaluate the rankings of these types by taking a weighted average of the probabilities obtained from both modules. Formally, we have
\begin{equation}~\label{reranking}
\resizebox{.86\linewidth}{!}{$
    \text{z}_i^j = 
    \begin{cases}
        \alpha \cdot p_i^j + (1-\alpha)\cdot q_i^j & \text{Rank}(q_i^j) \leq k\\
        q_i^j & \text{otherwise}
    \end{cases}
    , $}
\end{equation}
where $0\leq \alpha \leq 1$ is the hyperparameter that controls the balance between the two modules, $z_i^j$ is the aggregated probability that entity $i$ belongs to type $j$. These probabilities are ranked in descending order for model evaluation. Note that the aggregated probabilities will not be utilized to optimize the SEM or SKA modules. Such an \textit{unsupervised setting} prevents the final type predictions from being affected by false-negative samples.

\section{Experiments}\label{sec5}
\begin{table*}[ht]\centering
\small
\resizebox{0.95\linewidth}{!}{
\begin{tabular}{ccccccccccc}
\toprule
\multicolumn{1}{c|}{\multirow{2}{*}{\textbf{Methods}}}& \multicolumn{5}{c|}{\textbf{FB15kET}} & \multicolumn{5}{c}{\textbf{YAGO43kET}}\\
\multicolumn{1}{c|}{}& \multicolumn{1}{c}{\textbf{Hit@1}} & \multicolumn{1}{c}{\textbf{Hit@3}} & \multicolumn{1}{c}{\textbf{Hit@10}} & \multicolumn{1}{c}{\textbf{MR}} & \multicolumn{1}{c|}{\textbf{MRR}} & \multicolumn{1}{c}{\textbf{Hit@1}} & \multicolumn{1}{c}{\textbf{Hit@3}} & \multicolumn{1}{c}{\textbf{Hit@10}} & \multicolumn{1}{c}{\textbf{MR}} & \multicolumn{1}{c}{\textbf{MRR}} \\
\midrule
\multicolumn{11}{c}{\textit{Embedding-based methods}} \\
\multicolumn{1}{c|}{TransE} & 0.504 & 0.686 & 0.835 & 18 & 0.618 & \multicolumn{1}{|c}{0.304} & 0.497 & 0.663 & 393 &\multicolumn{1}{c}{0.427} \\
\multicolumn{1}{c|}{ComplEx} & 0.463 & 0.680 & 0.841 & 20 & 0.595 & \multicolumn{1}{|c}{0.316} & 0.504 & 0.658 & 631 &\multicolumn{1}{c}{0.435} \\
\multicolumn{1}{c|}{RotatE} & 0.523 & 0.699 & 0.840 & 18 & 0.632 & \multicolumn{1}{|c}{0.339} & 0.537 & 0.695 & 316 &\multicolumn{1}{c}{0.462} \\
\multicolumn{1}{c|}{CompoundE} & 0.525 & 0.719 & 0.859 & - & 0.640 & \multicolumn{1}{|c}{0.364} & 0.558 & 0.703 & - &\multicolumn{1}{c}{0.480} \\
\multicolumn{1}{c|}{SimKGC} & 0.210 & 0.348 & 0.545 & 46 & 0.317 & \multicolumn{1}{|c}{0.097} & 0.184 & 0.317 & 259 &\multicolumn{1}{c}{0.172} \\
\multicolumn{1}{c|}{ETE} & 0.385 & 0.553 & 0.719 & - & 0.500 & \multicolumn{1}{|c}{0.137} & 0.263 & 0.422 & - &\multicolumn{1}{c}{0.230} \\
\multicolumn{1}{c|}{ConnectE} & 0.496 & 0.643 & 0.799 & 42 & 0.590 & \multicolumn{1}{|c}{0.160} & 0.309 & 0.479 & - &\multicolumn{1}{c}{0.280} \\
\midrule
\multicolumn{11}{c}{\textit{Graph Neural Network-based methods}} \\
\multicolumn{1}{c|}{MRGAT} & 0.562 & 0.663 & 0.804 & - & 0.630 & \multicolumn{1}{|c}{0.243} & 0.343 & 0.482 & - &\multicolumn{1}{c}{0.320} \\
\multicolumn{1}{c|}{RACE2T} & 0.561 & 0.688 & 0.817 & - & 0.646 & \multicolumn{1}{|c}{0.248} & 0.376 & 0.523 & - &\multicolumn{1}{c}{0.344} \\
\multicolumn{1}{c|}{AttEt} & 0.517 & 0.677 & 0.821 & - & 0.620 & \multicolumn{1}{|c}{0.244} & 0.413 & 0.565 & - &\multicolumn{1}{c}{0.350} \\
\multicolumn{1}{c|}{RGCN} & 0.597 & 0.722 & 0.843 & 20 & 0.679 & \multicolumn{1}{|c}{0.281} & 0.409 & 0.549 & 397 &\multicolumn{1}{c}{0.372} \\
\multicolumn{1}{c|}{CET} & 0.613 & 0.745 & 0.856 & 19 & 0.697 & \multicolumn{1}{|c}{0.398} & 0.567 & 0.696 & 250 &\multicolumn{1}{c}{0.503} \\
\multicolumn{1}{c|}{MiNer} & 0.654 & 0.768 & 0.875 & 15 & 0.728 & \multicolumn{1}{|c}{0.412} & 0.589 & 0.714 & \underline{223} &\multicolumn{1}{c}{0.521} \\
\midrule
\multicolumn{11}{c}{\textit{Transformer-based methods}} \\
\multicolumn{1}{c|}{TET} & 0.638 & 0.762 & 0.872 & - & 0.717 & \multicolumn{1}{|c}{0.408} & 0.571 & 0.695 & - &\multicolumn{1}{c}{0.510} \\
\midrule
\multicolumn{1}{c|}{\textbf{SSET (w/ BERT)}} & \underline{0.693} & \underline{0.800} & \underline{0.895} & \underline{12} & \underline{0.761} & \multicolumn{1}{|c}{\underline{0.473}} & \underline{0.644} & \textbf{0.762} & 244 &\multicolumn{1}{c}{\underline{0.576}} \\
\multicolumn{1}{c|}{\textbf{SSET (w/ RoBERTa)}} & \textbf{0.697} & \textbf{0.804} & \textbf{0.898} & \textbf{11} & \textbf{0.765} & \multicolumn{1}{|c}{\textbf{0.476}} & \textbf{0.647} & \textbf{0.762} & \textbf{212} &\multicolumn{1}{c}{\textbf{0.579}} \\
\bottomrule
\end{tabular}
}
\caption{Experiment Results of KGET on FB15kET and YAGO43kET datasets. The best results are in \textbf{bold} and the second-best ones are \underline{underlined}. All results of baseline methods are referred from corresponding original papers except SimKGC. For SimKGC, we generate the results with the publicly available implementation. }
\label{result}
\vspace{-0.05cm}
\end{table*}

\subsection{Datasets}
We evaluate our model on two real-world KGs: FB15k~\cite{TransE} and YAGO43k~\cite{ETE}, which are derived from Google Freebase~\cite{Freebase} and YAGO knowledge base~\cite{YAGO}, respectively. Two entity typing datasets FB15kET~\cite{FB15kET} and YAGO43kET~\cite{ETE} provide entity-type assertions by mapping entities from the two KGs to their entity types. Regarding the textual information, we use the labels and descriptions of FB15k entities released by~\citet{Xie2016}. The YAGO43k dataset provides textual labels for each entity. We use Wikidata API to collect descriptions of entities in YAGO43k since the YAGO knowledge base is built upon Wikidata~\cite{Wikidata}.\footnote{The YAGO knowledge base assigns a distinct Wikidata ID to each of its entities, enabling us to access relevant information such as descriptions from the Wikidata API. } Relations and types in the two datasets are represented in textual identifiers. The statistics of datasets are shown in Appendix~\ref{dataset_appendix}.

\begin{table*}[ht]\centering
\resizebox{\linewidth}{!}{
\begin{tabular}{cccccccccccc}
\toprule
\multicolumn{1}{c|}{\multirow{2}{*}{\textbf{Exp.}}} & \multicolumn{1}{c|}{\multirow{2}{*}{\textbf{Model Settings}}}& \multicolumn{5}{c|}{\textbf{FB15kET}} & \multicolumn{5}{c}{\textbf{YAGO43kET}}\\
\multicolumn{1}{c|}{} & \multicolumn{1}{c|}{\textbf{}} & \multicolumn{1}{c}{\textbf{Hit@1}} & \multicolumn{1}{c}{\textbf{Hit@3}} & \multicolumn{1}{c}{\textbf{Hit@10}} & \multicolumn{1}{c}{\textbf{MR}} & \multicolumn{1}{c|}{\textbf{MRR}} & \multicolumn{1}{c}{\textbf{Hit@1}} & \multicolumn{1}{c}{\textbf{Hit@3}} & \multicolumn{1}{c}{\textbf{Hit@10}} & \multicolumn{1}{c}{\textbf{MR}} & \multicolumn{1}{c}{\textbf{MRR}} \\
\midrule






\multicolumn{1}{c|}{\textbf{1}} & \multicolumn{1}{c|}{SEM (MET) only} & 0.688 & 0.797 & 0.894 & 12 & 0.758 & \multicolumn{1}{|c}{0.468} & 0.636 & 0.756 & \textbf{162} &\multicolumn{1}{c}{0.571} \\

\multicolumn{1}{c|}{\textbf{2}} & \multicolumn{1}{c|}{SKA (w/o textual emb. \& KD)} & 0.671 & 0.778 & 0.878 & 15 & 0.741 & \multicolumn{1}{|c}{0.457} & 0.614 & 0.730 & 332 &\multicolumn{1}{c}{0.555} \\

\multicolumn{1}{c|}{\textbf{3}} & \multicolumn{1}{c|}{pre-trained RoBERTa + SKA (w/o KD)} & 0.665 & 0.780 & 0.881 & 15 & 0.738 & \multicolumn{1}{|c}{0.451} & 0.618 & 0.732 & 344 &\multicolumn{1}{c}{0.553} \\

\multicolumn{1}{c|}{\textbf{4}} &  \multicolumn{1}{c|}{SEM + SKA (w/o KD)} & 0.677 & 0.787 & 0.886 & 13 & 0.748 & \multicolumn{1}{|c}{0.458} & 0.622 & 0.740 & 265 &\multicolumn{1}{c}{0.558} \\

\multicolumn{1}{c|}{\textbf{5}} & \multicolumn{1}{c|}{SEM + SKA (w/ KD)} & 0.678 & 0.792 & 0.892 & 12 & 0.750 & \multicolumn{1}{|c}{0.461} & 0.633 & 0.751 & 213 &\multicolumn{1}{c}{0.566} \\

\multicolumn{1}{c|}{\textbf{6}} & \multicolumn{1}{c|}{SEM + SKA (w/ KD) + UTR} & \textbf{0.697} & \textbf{0.804} & \textbf{0.898} & \textbf{11} & \textbf{0.765} & \multicolumn{1}{|c}{\textbf{0.476}} & \textbf{0.647} & \textbf{0.762} & 212 &\multicolumn{1}{c}{\textbf{0.579}} \\
\bottomrule
\end{tabular}
}
\caption{Results for ablation studies with RoBERTa-base as backbone language model. The best results are in \textbf{bold}.}
\label{ablation}
\vspace{-0.15cm}
\end{table*}

\begin{table*}[ht]\centering
\footnotesize
\resizebox{\linewidth}{!}{
\begin{tabular}{cccccccccccc}
\toprule
\multicolumn{1}{c|}{\multirow{2}{*}{\textbf{Methods}}} & \multicolumn{1}{c|}{{\multirow{2}{*}{\textbf{$k$-hop}}}}& \multicolumn{5}{c|}{\textbf{FB15kET}} & \multicolumn{5}{c}{\textbf{YAGO43kET}}\\
\multicolumn{1}{c|}{} & 
\multicolumn{1}{c|}{ }&  \multicolumn{1}{c}{\textbf{Hit@1}} & \multicolumn{1}{c}{\textbf{Hit@3}} & \multicolumn{1}{c}{\textbf{Hit@10}} & \multicolumn{1}{c}{\textbf{MR}} & \multicolumn{1}{c|}{\textbf{MRR}} & \multicolumn{1}{c}{\textbf{Hit@1}} & \multicolumn{1}{c}{\textbf{Hit@3}} & \multicolumn{1}{c}{\textbf{Hit@10}} & \multicolumn{1}{c}{\textbf{MR}} & \multicolumn{1}{c}{\textbf{MRR}} \\
\midrule

\multicolumn{1}{c|}{RGCN} & \multicolumn{1}{c|}{$1$-hop} & 0.597 & 0.722 & 0.843 & 20 & 0.679 & \multicolumn{1}{|c}{0.281} & 0.409 & 0.549 & 397 &\multicolumn{1}{c}{0.372} \\

\multicolumn{1}{c|}{RGCN} & \multicolumn{1}{c|}{$2$-hop} & 0.580 & 0.709 & 0.830 & 29 & 0.664 & \multicolumn{1}{|c}{0.243} & 0.343 & 0.482 & 587 &\multicolumn{1}{c}{0.360} \\

\multicolumn{1}{c|}{MiNer} & \multicolumn{1}{c|}{$1$-hop} & 0.637 & 0.761 & 0.865 & 18 & 0.716 & \multicolumn{1}{|c}{0.402} & 0.580 & 0.710 & 245 &\multicolumn{1}{c}{0.512} \\

\multicolumn{1}{c|}{MiNer} & \multicolumn{1}{c|}{$2$-hop} & 0.653 & 0.764 & 0.871 & 15 & 0.726 & \multicolumn{1}{|c}{0.412} & 0.589 & 0.714 & 223 &\multicolumn{1}{c}{0.521} \\

\multicolumn{1}{c|}{MiNer} & \multicolumn{1}{c|}{$3$-hop} & 0.651 & 0.766 & 0.872 & 15 & 0.726 & \multicolumn{1}{|c}{0.411} & 0.589 & 0.713 & 224 &\multicolumn{1}{c}{0.520} \\

\multicolumn{1}{c|}{SSET (w/ RoBERTa)} & \multicolumn{1}{c|}{$1$-hop} & 0.696 & 0.802 & 0.896 & \textbf{11} & 0.764 & \multicolumn{1}{|c}{0.473} & 0.645 & 0.761 & 213 &\multicolumn{1}{c}{0.576} \\

\multicolumn{1}{c|}{SSET (w/ RoBERTa)} & \multicolumn{1}{c|}{$2$-hop} & \textbf{0.697} & \textbf{0.804} & \textbf{0.898} & \textbf{11} & \textbf{0.765} & \multicolumn{1}{|c}{0.475} & \textbf{0.647} & \textbf{0.762} & \textbf{206} &\multicolumn{1}{c}{0.578} \\

\multicolumn{1}{c|}{SSET (w/ RoBERTa)} & \multicolumn{1}{c|}{$3$-hop} & 0.696 & 0.803 & \textbf{0.898} & 12 & 0.764 & \multicolumn{1}{|c}{\textbf{0.476}} & \textbf{0.647} & \textbf{0.762} & 212 &\multicolumn{1}{c}{\textbf{0.579}} \\

\bottomrule
\end{tabular}
}
\caption{Results for ablation studies with different scopes of the local neighborhood. The best results are in \textbf{bold}. }
\label{multihopexp}
\vspace{-0.05cm}
\end{table*}

\subsection{Baselines}
In this section, we compare our proposed SSET model with $14$ baselines. Specifically, we consider traditional KG embedding methods TransE~\cite{TransE}, ComplEx~\cite{ComplEx}, RotatE~\cite{RotatE}, and CompoundE~\cite{CompoundE}; text-based KG completion method SimKGC~\cite{SimKGC}, embedding-based KGET methods ETE~\cite{ETE} and ConnectE~\cite{ConnectE}; GNN-based methods ConnectE-MRGAT~\cite{MRGAT}, RACE2T~\cite{RACE2T}, RGCN~\cite{RGCN}, AttEt~\cite{AttEt}, CET~\cite{CET}, and MiNer~\cite{Miner}; as well as transformer-based method TET~\cite{TET}.

\subsection{Main Results}
Table~\ref{result} summarizes the performance of SSET and all baselines on the two datasets. The experiment results show that the proposed model SSET significantly and consistently outperforms all baseline methods. Particularly, compared to the previous state-of-the-art method MiNer~\cite{Miner}, SSET achieves an absolute Hit@$1$ improvement of $4.3\%$ and $6.4\%$ on the FB15kET and the YAGO43kET datasets with a RoBERT-base textual encoder. It demonstrates that SSET is highly effective for the KGET task. 

Traditional KGE methods, such as TransE, ComplEx, RotatE, and CompoundE, cannot achieve desirable performance as they consider KGET as a link prediction task, which assumes that entities and their types are connected with a special relation ``\textit{has type}''. Due to the significant differences in textual semantics between entities and types, the text-based KGC model SimKGC lags far behind in the KGET task. GNN-based methods outperform embedding-based KGET methods ETE and ConnectE as GNNs can better capture the structural knowledge of entities within the local subgraph. It should be noted that RGCN, CET and MiNer perform better than other GNN-based methods since they consider KGET as a multi-label node classification task~\cite{multilabel}. This setting allows the model to mark all types that are not annotated to the target entity as negative samples, which typically improves inference performance. 

We have evaluated two variants of SSET to assess the effects of selecting different backbone language models. The experiment results show that both BERT and RoBERTa can help SSET to achieve state-of-the-art performance. Furthermore, employing a larger language model (RoBERTa) leads to better performance. We also observe that the performance improvement on the YAGO43kET dataset is particularly remarkable compared to the FB15kET dataset. This can be attributed to the sparser graph structure of YAGO43kET, where entities have significantly fewer neighbors ($7.8$ on avg.) than FB15kET ($32.3$ on avg.). In such cases, leveraging textual semantics of entities, relations, and types becomes crucially important.

In addition, the performance gain on Hit@$1$ is more pronounced compared to Hit@$3$ and Hit@$10$ on both datasets, highlighting the superior ability of our model to identify the most suitable type for a given entity. This can be explained by the incorporation of the unsupervised type re-ranking module, which ensures that the top-ranked types are simultaneously supported by both semantic and structural knowledge. 
We will provide relevant examples in the case study section.

\begin{table*}[ht]\centering
\resizebox{\linewidth}{!}{
\begin{tabular}{ccccccccc}
\toprule

\multicolumn{1}{c}{\multirow{2}{*}{\textbf{Entity}}} & \multicolumn{1}{c|}{\textbf{Missing}} & \multicolumn{1}{c|}{\multirow{2}{*}{\textbf{Module}}} & \multicolumn{2}{c|}{\textbf{Rank \#1}} & \multicolumn{2}{c|}{\textbf{Rank \#2}} & \multicolumn{2}{c}{\textbf{Rank \#3}} \\

\multicolumn{1}{c}{} & \multicolumn{1}{c|}{\textbf{Type}} & \multicolumn{1}{c|}{} & \multicolumn{1}{c}{\textbf{Type}} & \multicolumn{1}{c|}{\textbf{Score}} & \multicolumn{1}{c}{\textbf{Type}} & \multicolumn{1}{c|}{\textbf{Score}} & \multicolumn{1}{c}{\textbf{Type}} & \multicolumn{1}{c}{\textbf{Score}} \\

\midrule
\multicolumn{1}{c}{\multirow{3}{*}{\makecell{/m/0k8z \\ (Apple Inc.)}}} & \multicolumn{1}{c|}{\multirow{3}{*}{\textcolor{Bittersweet}{/popstra/company}}} & \multicolumn{1}{c|}{SEM} & /business/brand
& \multicolumn{1}{c|}{0.947} & \textcolor{Bittersweet}{/popstra/company} & \multicolumn{1}{c|}{0.927} & /business/sponsor & \multicolumn{1}{c}{0.888} \\

\multicolumn{1}{c}{} & \multicolumn{1}{c|}{} & \multicolumn{1}{c|}{SKA} & /ontologies/ontology\_instance & \multicolumn{1}{c|}{0.965} & \textcolor{Bittersweet}{/popstra/company} & \multicolumn{1}{c|}{0.958} & /fblinux/topic & \multicolumn{1}{c}{0.904} \\

\multicolumn{1}{c}{} & \multicolumn{1}{c|}{} & \multicolumn{1}{c|}{UTR} & \textcolor{Bittersweet}{/popstra/company} & \multicolumn{1}{c|}{0.942} & /ontologies/ontology\_instance & \multicolumn{1}{c|}{0.900} & /business/sponsor & \multicolumn{1}{c}{0.876} \\

\midrule

\multicolumn{1}{c}{\multirow{3}{*}{\makecell{/m/02x\_h0 \\ (Jermaine Dupri)}}} & \multicolumn{1}{c|}{\multirow{3}{*}{\textcolor{ForestGreen}{/music/composer}}} & \multicolumn{1}{c|}{SEM} & /celebrities/celebrity 
& \multicolumn{1}{c|}{0.942} & \textcolor{ForestGreen}{/music/composer} & \multicolumn{1}{c|}{0.919} & /broadcast/artist & \multicolumn{1}{c}{0.859} \\

\multicolumn{1}{c}{} & \multicolumn{1}{c|}{} & \multicolumn{1}{c|}{SKA} & /broadcast/artist & \multicolumn{1}{c|}{0.970} & \textcolor{ForestGreen}{/music/composer} & \multicolumn{1}{c|}{0.969} & /internet/social\_network\_user & \multicolumn{1}{c}{0.950} \\

\multicolumn{1}{c}{} & \multicolumn{1}{c|}{} & \multicolumn{1}{c|}{UTR} & \textcolor{ForestGreen}{/music/composer} & \multicolumn{1}{c|}{0.944} & /broadcast/artist & \multicolumn{1}{c|}{0.915} & /internet/social\_network\_user & \multicolumn{1}{c}{0.832} \\

\midrule

\multicolumn{1}{c}{\multirow{3}{*}{\makecell{/m/0ptk\_ \\ (Canadian Dollar)}}} & \multicolumn{1}{c|}{\multirow{3}{*}{\textcolor{NavyBlue}{/finance/currency}}} & \multicolumn{1}{c|}{SEM} & \textcolor{NavyBlue}{/finance/currency}
& \multicolumn{1}{c|}{0.955} & /military/military\_post & \multicolumn{1}{c|}{0.490} & /location/location & \multicolumn{1}{c}{0.414} \\

\multicolumn{1}{c}{} & \multicolumn{1}{c|}{} & \multicolumn{1}{c|}{SKA} & /sports/sports\_team\_location & \multicolumn{1}{c|}{0.974} & \textcolor{NavyBlue}{/finance/currency} & \multicolumn{1}{c|}{0.971} & /award/award\_nominee & \multicolumn{1}{c}{0.921} \\

\multicolumn{1}{c}{} & \multicolumn{1}{c|}{} & \multicolumn{1}{c|}{UTR} & \textcolor{NavyBlue}{/finance/currency} & \multicolumn{1}{c|}{0.963} & /location/location & \multicolumn{1}{c|}{0.536} & /military/military\_post & \multicolumn{1}{c}{0.524} \\
\bottomrule
\end{tabular}
}
\caption{Representative type re-ranking examples. We present top 3 type candidates and corresponding relevance scores from each of the three modules in SSET. }
\label{case}
\vspace{-0.4cm}
\end{table*}

\subsection{Ablation Studies}
We verify the effectiveness of each component in the SSET model by answering the following research questions~(RQ). Table~\ref{ablation} shows the experiment results for ablation studies.
\paragraph{RQ1: Can the SEM and the SKA module generate plausible inference results?} In \textbf{Exp.~1}, we directly utilize the MET fine-tuning task proposed in the SEM module to predict missing types. From the experiment results, we observe that the SEM module significantly and consistently outperforms all baseline methods across all metrics. With semantic knowledge from the textual representations in KGs, the SEM module only requires minimal structural knowledge ($1$-hop triples and known types of the target entity) to achieve significant performance improvement. In contrast, the GNN-based method MiNer~\cite{Miner} needs to consider $4$-hop neighbors of each entity to achieve its best performance on FB15kET. This shows that the presence of textual semantics can compensate for the absence of multi-hop neighbors.

In \textbf{Exp. 2}, we utilize the SKA module to perform KGET. The embeddings of entities, relations, and types are randomly initialized and will be optimized as model parameters. Without additional knowledge from the textual representations, the proposed SKA module still significantly outperforms all baseline methods.
\paragraph{RQ2: Does language model fine-tuning improve the model performance?} 
In \textbf{Exp.~3} and \textbf{Exp.~4}, we separately employ the pre-trained RoBERTa model and the fine-tuned RoBERTa model (obtained from the SEM module) to embed textual information of entities, relations, and types.  The performance gap between \textbf{Exp.~2} and \textbf{Exp.~4} indicates that the textual embeddings from the fine-tuned RoBERTa model provide crucial semantic knowledge for type inference. However, the expected performance improvement is not observed in \textbf{Exp.~3} since the pre-trained RoBERTa model lacks factual knowledge from the KG, which confirms the effectiveness and necessity of the SEM module.
\paragraph{RQ3: Does knowledge distillation alleviate the false-negative problem?} Among all metrics, the Mean Rank (MR) is more susceptible to the false negative problem since types with a high rank significantly increase the MR value. In \textbf{Exp.~5}, we incorporate the knowledge distillation~(KD) loss into the training objective of the SKA module. Comparing the results between \textbf{Exp.~4} and \textbf{Exp.~5}, 
the MR exhibits a significant decrease (7.7\% on FB15kET and 19.6\% on YAGO43kET), empirically confirming that the KD loss can mitigate the impact of readily identifiable false-negative samples.
\paragraph{RQ4: Can unsupervised type re-ranking improve the entity typing performance?} 
Comparing the results between \textbf{Exp.~5} and \textbf{Exp.~6}, we observe a remarkable performance improvement with the inclusion of the unsupervised type re-ranking module. In contrast, relying solely on either SEM or SKA module leads to suboptimal results, which suggests that both transformers and GNNs have certain limitations when it comes to modeling knowledge within a KG. Notably, despite not incurring any additional computational overhead during training, the incorporation of the re-ranking module results in a more significant performance gain compared to using textual embeddings or employing the knowledge distillation loss. This observation highlights that re-ranking is a direct, effective, and efficient knowledge integration approach. 

\paragraph{RQ5: How do multi-hop neighbors affect the performance of SSET?} 
From Table~\ref{multihopexp} we observe that leveraging multi-hop neighbors generally improves the performance of the proposed SSET model.
Specifically, including $2$-hop neighbors is sufficient for SSET to attain optimal results on the FB15kET dataset. In contrast, to achieve the best performance on the YAGO43kET dataset, SSET needs to aggregate knowledge from 3-hop neighbors. This observation can also be justified by the sparser graph structure of the YAGO43kET dataset. 
Furthermore, the proposed SSET model surpasses GNN-based methods RGCN~\cite{RGCN} and MiNer~\cite{Miner} with neighbor entities in different numbers of hops, demostrating that the proposed SSET model is robust to the scope of structural knowledge. 

\subsection{Case Study}
In Table~\ref{case}, we selected three representative inference results obtained from the three modules of the SSET model. 
These examples show how type re-ranking improves the accuracy of type inference. In the first example, the missing type \textit{/popstra/company} of entity \textit{/m/0k8z} (\textit{Apple Inc.}) is initially considered as a negative sample, resulting in a limited relevance score during the training process. However, other candidate types, namely \textit{/business/brand} and \textit{/ontologies/ontology\_instance}, fail to obtain sufficient support from both the SEM and the SKA modules. Consequently, after re-ranking, we can conclude that \textit{/popstra/company} is the most appropriate type for entity \textit{/m/0k8z}. The second example follows a similar pattern. 

In the third example, the SEM module determines that entity \textit{/m/0ptk} (\textit{Canadian Dollar}) belongs to the missing type \textit{/finance/currency} based on the textual semantics provided in the local neighborhood.~\footnote{To mitigate overfitting and prevent potential information leakage, we refrain from utilizing the textual semantics of the target entity in the process of type inference.} In contrast, the SKA module is unable to clearly differentiate between this type and other distractors. Nevertheless, type re-ranking can help the model to make desirable inferences. 
It further confirms the critical role of integrating textual semantics is in the KGET task. 

\section{Conclusion} 
In this paper, we propose a novel model SSET for the KGET task. SSET focuses on modeling the integration and interaction between semantic and structural knowledge from various perspectives. The SEM module utilizes structural triples and entity-type assertions as the corpus to fine-tune the PLM. The SKA module fuses pre-trained textual embeddings into learnable structural embeddings to address the ``cold start'' issue and leverages the preliminary inference results from the SEM module to mitigate the false-negative problem. Finally, the UTR module enables us to combine inference scores from different modules with minimal computational overhead. The experiment results show that the proposed SSET model significantly and consistently outperforms all baseline methods.

\section*{Limitations}
Although our proposed method achieved a significant breakthrough on the KGET task, it still has some limitations to be addressed in the future. First, the proposed SSET model is not capable of handling ``Inductive Entity Typing''. The KGET task discussed in this paper, along with related studies, operates under a ``transductive setting''. The transductive setting focuses on inferring missing entity-type assertions with the set of types present in the KG. 
We plan to tackle unseen types that are not mentioned in the KG in the future.
In addition, the proposed FET model cannot deal with the fine-grained entity typing~{(FET)} task. The FET task and the KGET task differ in the objective, the former predicts the types for entity mentions within natural language sentences, while the latter focuses on entities in KGs. We also leave the FET task as a future work.

\section*{Acknowledgements}
Part of the research reported in this paper was done when Ho-fung Leung was with The Chinese University of Hong Kong. The research presented in this paper was partially supported by the Research Grants Council of the Hong Kong Special Administrative Region, China (CUHK 14222922, RGC GRF 2151185).

\bibliography{anthology,custom}

\clearpage
\appendix
\section{Re-weighting Function in the SFNA Loss}
In the SFNA Loss~\cite{TET}, the reweighting function $f(\cdot)$ will assign a lower weight to the negative samples with too high or too low probabilities. The negative samples with a high type prediction probability are very likely to be false-negative samples. The negative samples with a low type prediction probability are usually easy samples. The application of the following re-weighting function~\cite{TET} allows our model to pay less attention to these negative samples.  
\begin{equation}~\label{reweighting}
    f(x) = 
    \begin{cases}
        3x-2x^2, & x\leq 0.5 \\
        x-2x^2+1, & x>0.5
    \end{cases}
    .
\end{equation}


\section{The Statistics of Datasets}~\label{dataset_appendix}
The statistics of the FB15kET and YAGO43kET datasets are shown in Table~\ref{dataset}.
\begin{table}[htbp]\centering
\small
\begin{tabular}{ccc}
\toprule
\textbf{Dataset}& \textbf{FB15kET} & \textbf{YAGO43kET}\\
\midrule
\#Entities & 14,951 & 42,335 \\
\#Relations & 1,345 & 37 \\
\#Types & 3,851 & 45,182 \\
\#Triples & 483,142 & 331,687 \\
\#Train & 136,618 & 375,853 \\
\#Valid & 15,749 & 42,739 \\
\#Test & 15,780 & 42,750 \\
\bottomrule
\end{tabular}
\caption{Statistics of Datasets}
\label{dataset}
\vspace{-0.25cm}
\end{table}

\section{Evaluation Metrics}~\label{metrics}
For each test sample, we calculate the probabilities that the entity belongs to each of the types within the KG and then sort them in descending order. For a fair comparison, we adopt the filtered setting~\cite{TransE} to remove all type annotations existed in training, validation, and test sets before ranking. We compute Hits@$1$, Hits@$3$, Hits@$10$, Mean Rank (MR), and Mean Reciprocal Rank (MRR) for performance evaluation. Higher results stand for better performance for all evaluation metrics except MR.

\section{Experiment Settings}~\label{expsettings}
Our implementation of the semantic knowledge encoder is based on the ``BERT-base-uncased'' and the ``RoBERTa-base'' PLM from the Huggingface transformer library. We use Adam optimizer~\cite{Adam} to optimize the module parameters. The initial learning rate $1e\text{-}4$ remains unchanged within the initial decay interval of $50$ epochs. After that, we half the learning rate and double the decay interval. We fine-tune the BERT~\cite{BERT} and the RoBERTa~\cite{su-etal-2022-roberta} model with a batch size of $32$ for a maximum of $400$ epochs. During evaluation, we set the batch size to $1$ due to the GPU memory limitation. The tokenized sequences within each batch will be padded to the length of the longest sequence. We adopt similar experiment settings for the SKA module except that the learning rate $1e\text{-}3$ and $2e\text{-}3$ on the two datasets remains unchanged. For each entity, we sample $7$ triples and sample $3$ known types during training. All triples and known types of the target entity will be used in model inference. We select the best hyper-parameters based on the model performance on the validation set. Detailed parameter settings are listed in Table~\ref{hyperparameters}.

\begin{table}[htbp]\centering
\small
\begin{tabular}{ccc}
\toprule
\textbf{Dataset}& \textbf{FB15kET} & \textbf{YAGO43kET}\\
\midrule
SEM learning rate & $5e\text{-}4$ & $5e\text{-}4$ \\
SKA learning rate & $1e\text{-}3$ & $2e\text{-}3$ \\
Training batch size & $32$ & $32$ \\
Textual Emb. dim. & $768$ & $768$ \\
Structural Emb. dim. & $100$ & $100$ \\
\# triples sampled: $m$ & $7$ & $3$ \\
\# known types sampled: $n$ & $8$ & $3$ \\
\# Heads in CSRA: $H$ & $5$ & $5$ \\
$\lambda$ & $0.75$ & $0.8$ \\
$\alpha$ & $0.5$ & $0.5$ \\
\bottomrule
\end{tabular}
\caption{The hyperparameters of SSET model in the FB15kET and YAGO43kET datasets}
\label{hyperparameters}
\vspace{-0.25cm}
\end{table}

\begin{figure*}
    \centering
    \includegraphics[width=0.90\textwidth]{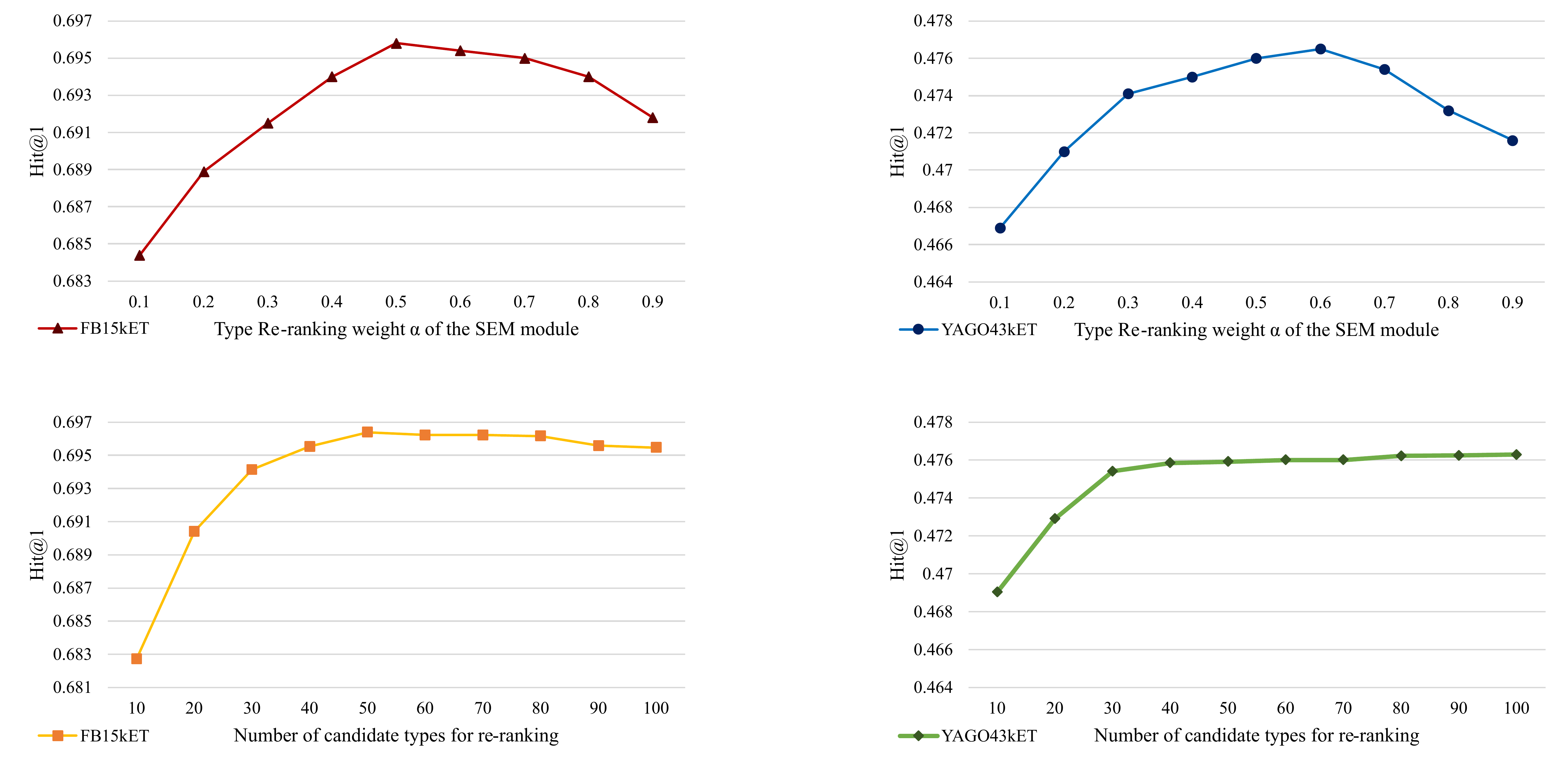}
    \caption{Experimental results of ablation studies with different experimental conditions. }
    \label{hyperparameters}
\vspace{-0.15cm}
\end{figure*}

\section{Complexity Analysis}
We evaluate the computational cost of the SEM and SKA modules based on the number of floating-point operations~(FLOPs) required during the training or inference stage. It should be noted that the computational cost of the SKA module depends on the scope of the knowledge aggregation. The results corresponding to the two datasets are listed in Table~\ref{flops}.
\begin{table}[htbp]\centering
\small
\begin{tabular}{ccc}
\toprule
\textbf{Module}& \textbf{FB15kET} & \textbf{YAGO43kET}\\
\midrule
SEM (w/ BERT) & 21.82G & 23.25G \\
SEM (w/ RoBERTa) & 25.28G & 25.63G \\
SKA (1-hop) & 327.6M & 2.15G \\
SKA (2-hop) & 1.74G & 5.27G \\
SKA (3-hop) & 10.99G & 20.09G \\
\bottomrule
\end{tabular}
\caption{The number of FLOPs required by the SEM and SKA modules on different datasets with different language models.}
\label{flops}
\vspace{-0.15cm}
\end{table}

All experiments with the BERT language model are conducted on a Lenovo workstation with an Intel Xeon W-2123 processor and an NVIDIA GeForce RTX3090 GPU with 24GB memory. We evaluated SSET with RoBERTa language model on a server with an Intel Xeon Platinum 8358 processor and 8 NVIDIA A100 40G GPUs.~\footnote{Only 1 GPU is used in our experiments.} We should emphasize that such configurations are not essential for our model to generate satisfactory entity typing results. The estimated times for modules of SSET to converge (or finish) are listed in Table~\ref{time}.
\begin{table}[htbp]\centering
\small
\begin{tabular}{ccc}
\toprule
\textbf{Module}& \textbf{FB15kET} & \textbf{YAGO43kET}\\
\midrule
SEM (w/ BERT) & 7 hours & 17.5 hours \\
SEM (w/ RoBERTa) & 7 hours & 19.7 hours \\ 
SKA (1-hop) & 1.3 hours & 6.1 hours \\
SKA (2-hop) & 1.5 hours & 7.3 hours \\
SKA (3-hop) & 4.3 hours & 21.4 hours \\
UTR (inference only) & < 10 seconds & 4 mins \\
\bottomrule
\end{tabular}
\caption{Estimated time for different modules in SSET to achieve its best results. }
\label{time}
\vspace{-0.10cm}
\end{table}

The statistics above show that the main computational overhead of our model arises from language model fine-tuning. Nevertheless, compared to other language model fine-tuning tasks~\cite{fastbert}, the computational cost of the SEM module remains reasonable. 
Despite this, considering the evident performance improvement, we maintain the belief that incorporating textual information in KGET is both necessary and feasible. In the future, we plan to explore the integration of efficient parameter tuning techniques (e.g., LoRA~\cite{LoRA}) in the domain of KGs. On the other hand, the computation cost of the SKA module exhibits a noticeable border effect. Expanding the scope of knowledge aggregation to more hops is unlikely to yield remarkable performance improvements, but will substantially increase the training time. 

\section{Additional Results with Different Hyperparameter Settings.}
\paragraph{Impact of Re-ranking Weights.} We conduct additional ablation experiments to analyze the effects of using different re-ranking weights $\alpha$ for module balancing. A higher value of $\alpha$ allows our model to prioritize semantic knowledge in the text, whereas a lower value of $\alpha$ directs our model to focus more on the structural knowledge of the target entity. From Figure~\ref{hyperparameters} (top) we can see that a moderate value of $\alpha$ (e.g. $0.5$ on the FB15kET dataset and $0.6$ on the YAGO43kET dataset) maximizes the performance of type re-ranking. This reaffirms the substantial complementary relationship between semantic and structural knowledge. 

\paragraph{Impact of Selecting Different Numbers of Type Candidates for Re-ranking. }
We further investigate the impact of using  different numbers of type candidates for re-ranking. From Figure~\ref{hyperparameters} (bottom) we can conclude that increasing the number of type candidates generally enhances the model performance. Specifically, selecting an insufficient number of candidates excludes plausible types supported by high probabilities derived from semantic knowledge. However, employing an excessive number of re-ranking candidates may degrade the model performance as structural knowledge cannot provide sufficient support for type candidates with relatively lower probabilities.

\end{document}